\documentclass{isprs} 
\usepackage{subfigure}
\usepackage{setspace}
\usepackage{geometry} 
\usepackage{epstopdf}
\usepackage[labelsep=period]{caption}  
\usepackage[british]{babel} 
\usepackage[hang]{footmisc}
\usepackage{amsmath}
\usepackage{natbib}
\usepackage{tikz}

\begin{document}

\title{Combining HoloLens with Instant-NeRFs: Advanced Real-time 3D Mobile Mapping}

\author{
Dennis Haitz\textsuperscript{1}\thanks{Corresponding author}, Boris Jutzi\textsuperscript{1}, Markus Ulrich\textsuperscript{1}, Miriam Jäger\textsuperscript{1}, Patrick Hübner\textsuperscript{1, 2} }

\address{
	\textsuperscript{1 } Institute for Photogrammetry and Remote Sensing, Karlsruhe Institute of Technology (KIT), Karlsruhe, Germany\\ 
    \{dennis.haitz, boris.jutzi, markus.ulrich, miriam.jaeger\}@kit.edu.\\
	\textsuperscript{2 } Remote Sensing and Image Analysis, Technical University of Darmstadt, Darmstadt, Germany\\
    patrick.huebner@tu-darmstadt.de\\
}


\abstract{This work represents a large step into modern ways of fast 3D reconstruction based on RGB camera images. Utilizing a Microsoft HoloLens 2 as a multisensor platform that includes an RGB camera and an inertial measurement unit for SLAM-based camera-pose determination, we train a Neural Radiance Field (NeRF) as a neural scene representation in real-time with the acquired data from the HoloLens. The HoloLens is connected via Wifi to a high-performance PC that is responsible for the training and 3D reconstruction. After the data stream ends, the training is stopped and the 3D reconstruction is initiated, which extracts a point cloud of the scene. With our specialized inference algorithm, five million scene points can be extracted within 1 second. In addition, the point cloud also includes radiometry per point. Our method of 3D reconstruction outperforms grid point sampling with NeRFs by multiple orders of magnitude and can be regarded as a complete real-time 3D reconstruction method in a mobile mapping setup.}

\maketitle

\sloppy

\section{Introduction}\label{sec:introduction}

The 3D reconstruction of industrial objects is a central task in the field of machine vision regarding quality control. Geometric representations can be used for the detection of geometric damage on the surface of such objects. These types of damages set the object's shape off of some geometric target state, which needs to be achieved or kept within an industrial production line in order to ensure certain quality constraints.
Typically, in industrial damage detection settings, it is important to achieve fast detection and thus 3D reconstruction in real-time or near real-time in order for detection steps not to become the bottleneck constraining conveyor belts or other processes to work slower then necessary.
Besides, on-the-fly 3D reconstruction also enables on-the-fly 3D visualization of detected anomalies which can be of great benefit in on-premise inspection tasks.
In this context, Virtual (VR) and Augmented Reality (AR) devices hold great potential for the in-situ visualization of results derived from the analysis of 3D sensed data.\\
A current, promising approach for 3D reconstruction are Neural Radiance Fields (NeRF), where 3D scene geometry is represented implicitly by the weights of a small, fully-connected neural network.
The seminal paper on NeRF from \citet{mildenhall2020nerf} immediately sparked an impressive surge in research efforts with several hundreds of publications in the two subsequent years \citep{gao_et_al_2022}, for which the term 'the NeRF explosion' has been coined \citep{dellart_yen-chen_2020}.
While early approaches took training times in the order of hours to days, current approaches like \citep{mueller2022instantNgp} enable training of NeRFs in a matter of seconds to few minutes.\\
Thus, NeRF also holds potential for applications in the context of industrial object inspection.
However, NeRF approaches typically rely on camera poses which can be derived by photogrammetric techniques such as Structure-from-Motion (SfM).
This, however, necessitates to have all training images and their poses available beforehand and thus contradicts the proposed scenario of on-the-fly real-time object reconstruction.
For the realization of such a scenario, it is necessary to adapt NeRF to work on a continuous stream of input images with poses.
These can be obtained by means of real-time SLAM systems such as for instance the HoloLens which was found to have a pose accuracy in the range of few centimeters and few degrees \citep{huebner_et_al_2020} and can be readily used out-of-the-box.
As demonstrated in \citep{jaeger_et_al_2023}, the accuracy of the HoloLens poses is sufficient for NeRF training to converge.\\
This work presents an acquisition pipeline for a real-time image and pose streaming through a TCP client-server application and practically simultaneous Instant-NeRF training. The HoloLens acts as an image and pose server, while the Instant-NeRF implementation is extended by a client application, that receives images and writes them into a GPU image buffer. During the streaming process, we train the Instant-NeRF on-the-fly incrementally on the incoming image data. 
Furthermore, a fast geometric reconstruction of the scene is applied by querying the trained network based on sample rays from the training poses.
Thus, when the user finishes data acquisition by walking around an object of interest and looking at it while wearing the HoloLens, a neural scene representation has already been trained on-the-fly during the acquisition process and an explicit scene representation in the form of a dense point cloud is readily available for inspection.

\section{Technical Background}\label{sec:relatedwork}
\begin{figure}
    \centering
    \includegraphics[width=0.25\textwidth]{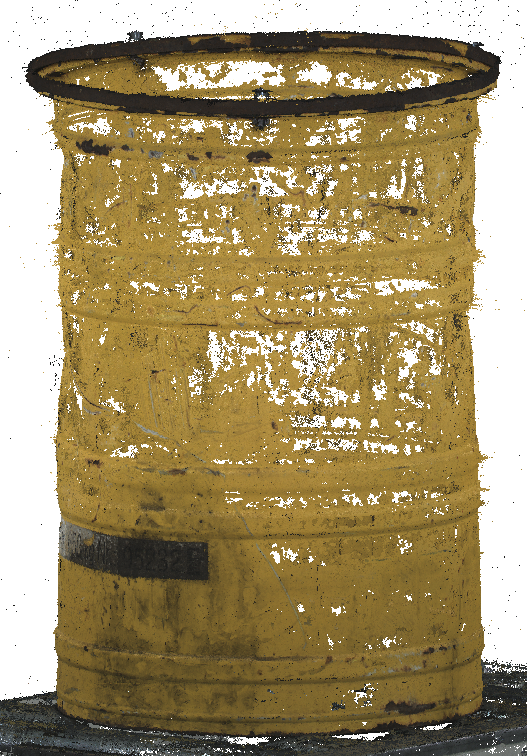}
    \caption{A point cloud extracted from multiple images of the barrel using Multi-view Stereo.}
    \label{fig:barrelmvs}
\end{figure}

\begin{figure}[h!]
  \centering
  \begin{minipage}[b]{0.45\linewidth}
    \includegraphics[height=60mm]{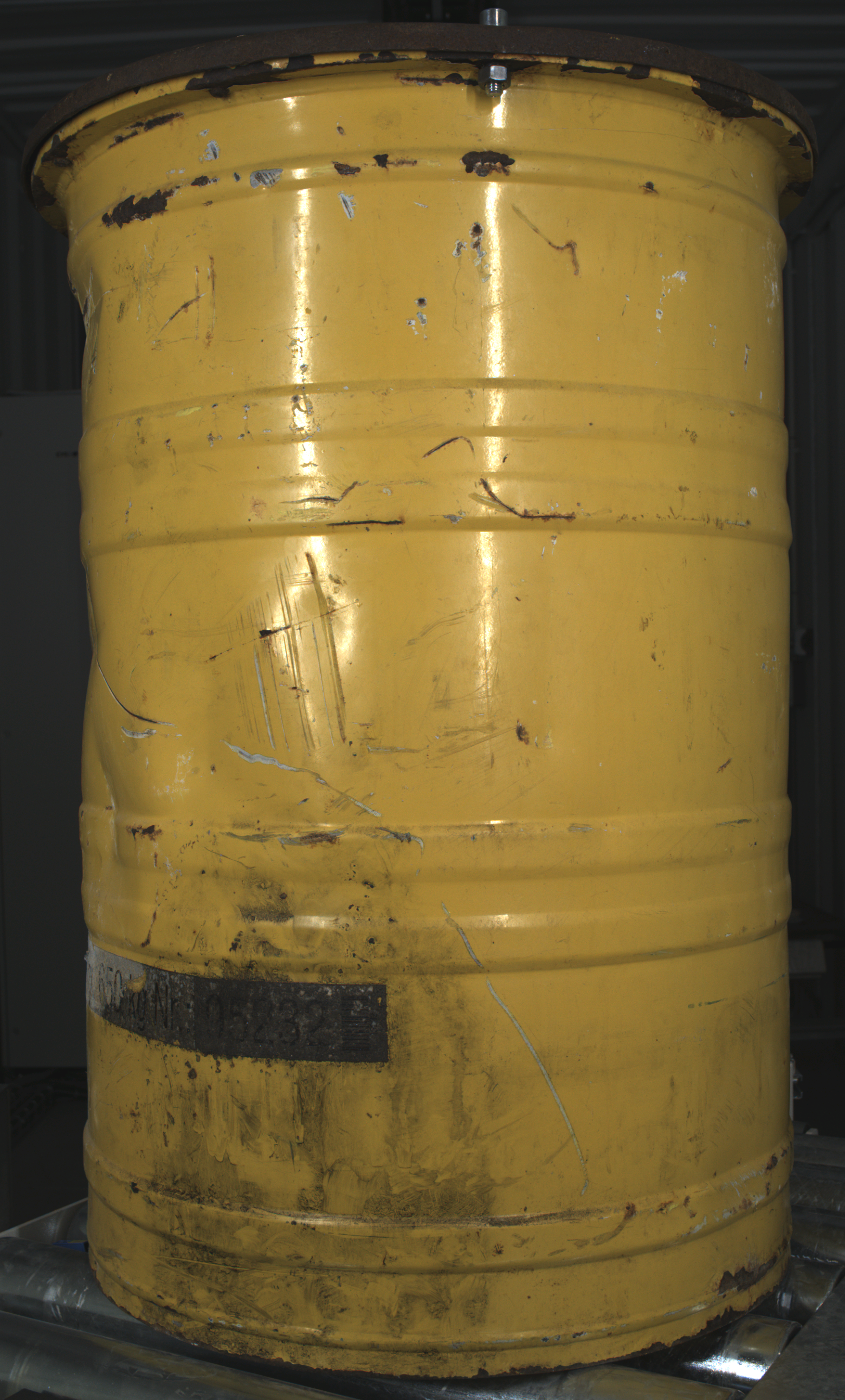}
    \caption*{(a)}
  \end{minipage}
  \begin{minipage}[b]{0.45\linewidth}
    \includegraphics[height=60mm]{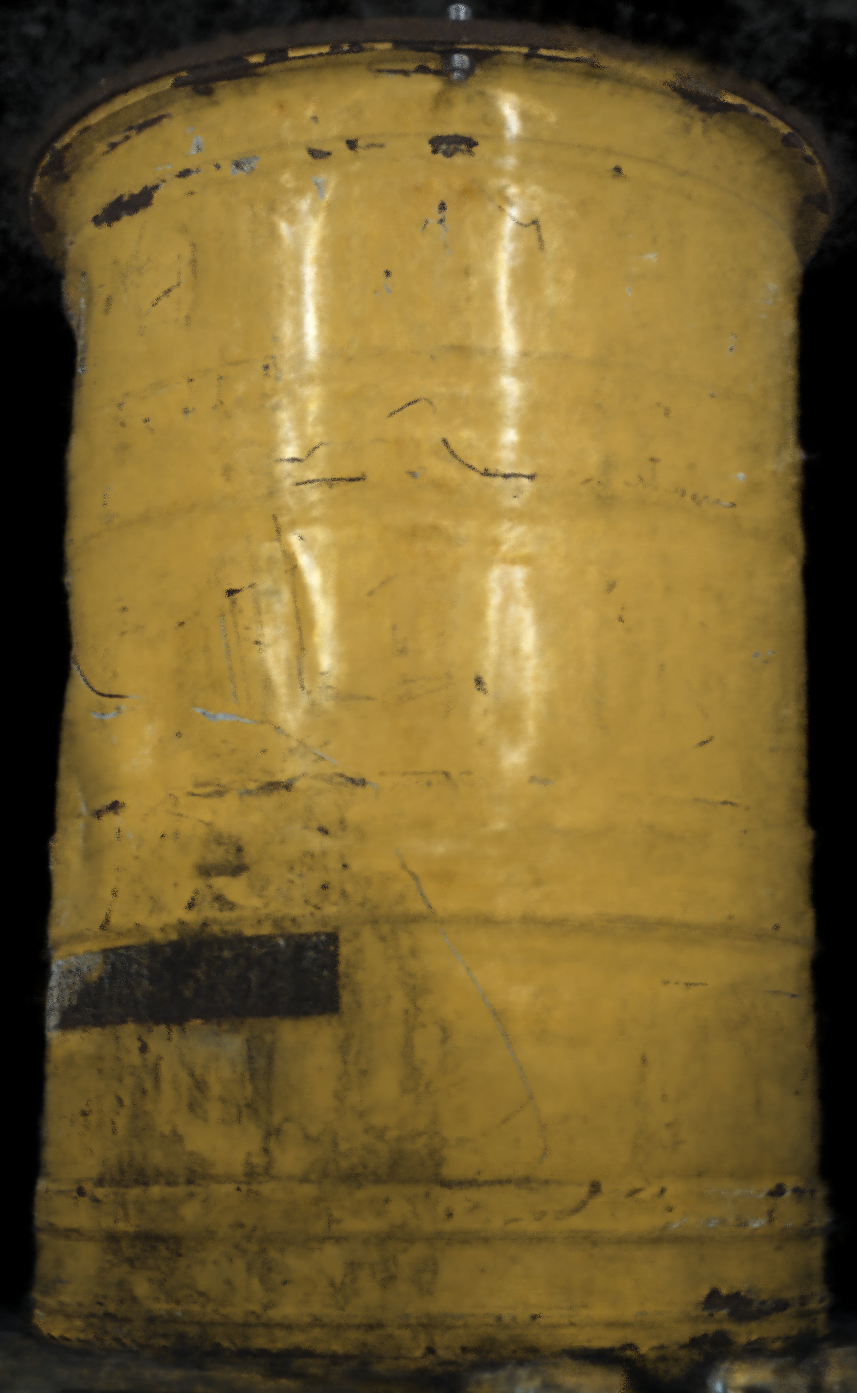}
    \caption*{(b)}
  \end{minipage}
  \caption{Image of the object to be reconstructed (a) and a NeRF-rendered image (b). To be more specific, the rendering is done via the Instant-NeRF implementation. The rendered image is from a similar, but nevertheless novel view w.r.t. the shown image in (a). On the left side of the rendered image, fairly large dents (geometric damage) of the object are visible through contour indentations.}
  \label{fig:barrel_comparison}
\end{figure}

Besides active sensors, e.g. based on triangulation, time-of-flight or phase difference measurements \citep{JutziMeyerHinz2015_1000070460, dubiosjutzi2021}, passive sensors like RGB cameras can be utilized to obtain 3D object models by applying photogrammetric methods. For a stereo image capture, at least two images need to be acquired, covering the same scene from different viewpoints. The interior and relative orientation need to be known or determined for the 3D reconstruction. The interior orientation can be obtained through a dedicated camera calibration and therefore needs to be done before the image acquisition. For fixed setups, the relative orientation can be obtained in advance by a camera calibration as well \citep{StegerUlrichWiedemann2018}. For varying setups, it can be obtained directly from the images that were acquired from the object by computing the fundamental matrix or the essential matrix, respectively \citep{Hartley2004}.
Furthermore, for obtaining the exterior orientation, a bundle adjustment can be applied. Usually this is not needed for obtaining a 3D model, however, if the exterior orientation is known in advance, e.g. through inertial measurement units (IMUs), the relative orientation can be derived trivially. A bundle adjustment can also be utilized to estimate the parameters of the interior orientation, whereupon a proper camera calibration might lead to a more accurate determination of the interior orientation. Structure-from-Motion (SfM) is a sparse reconstruction method based on feature matching for obtaining a 3D object model from two or more images of a scene. SfM estimates the interior, exterior and relative orientation similar to the aforementioned photogrammetric methods. Multi-view Stereo (MVS) is a dense reconstruction method that extends the described methods. It is used to obtain depth for every pixel in the set of overlapping images (Figure \ref{fig:barrelmvs}). The main disadvantage of this method is an extensive computation time.\\
Industrial production settings usually consist of stationary sensor systems and moving objects, e.g. on conveyor belts or rotary tables. But there are also special cases in which an object is stationary and the sensor moves around the object, e.g. on a robot arm or through a human operator.
The data processing for obtaining a 3D object model can be realized by means of the above mentioned photogrammetric techniques. In recent years, however, a new method representing 3D scenes through implicit fields emerged, as well as a large body of research that still grows continuously. This new method is called Neural Radiance Fields (NeRFs), which has its origin in the computer graphics community \citep{mildenhall2020nerf}. The objective of a NeRF is to render new views from a set of images (Figure \ref{fig:barrel_comparison}a) with their corresponding interior and exterior orientation. Throughout the rest of this work, we refer to interior and exterior orientation as camera parameters and (camera) pose, respectively. A NeRF learns properties of a scene, namely density \begin{math}\sigma\end{math} and a radiance \begin{math}\textbf{c}\end{math}, by optimizing a neural network. This neural network consists of multiple linear layers and activation layers in the original implementation, which qualifies it as a Multilayer Perceptron (MLP), a term that is widely used in the NeRF literature and was primarily used for describing multi-layer neural networks in the pre-deep-learning era \citep{mlp-rosenblatt1961principles}. A so-called volume rendering integrates a set of functions of \begin{math}\sigma\end{math} and \begin{math}\textbf{c}\end{math} over a ray, constructed from the projection center through the scene at first with equidistant sampling, and in a second, more detailed manner, according to a probability density function, centered at the highest density response from the former sampling. The volume rendering yields a color representation \begin{math}\hat{C}\end{math} at this particular position in the rendered image (Figure \ref{fig:barrel_comparison}b). After the execution of a \begin{math}L^2\end{math}-based loss function that takes the rendered image and the ground truth image, a backward pass is executed. The radiance \begin{math}\textbf{c}\end{math} is represented by a three dimensional vector that describes the emitted light on a surface and can be interpreted as a color information that is dependent on the viewing direction. The density \begin{math}\sigma\end{math} in the scene is a direction-independent property, which can be utilized to assume object geometry at the queried position. A density can be thought of as the probability for a point existing in scene space, belonging to an actual object.\\
While training and inference of NeRFs initially was a relatively slow process where training took multiple hours for a small set of 30 to 50 training images and poses, Instant-NGP \citep{mueller2022instantNgp} is a framework that provides strategies that lead to a remarkable acceleration regarding training and inference time. While Instant-NGP is the base framework for a few methods such as Signed Distance Fields (SDF), specific implementation developments for NeRFs have been applied \citep{muellerInstantNerf}, e.g. pose refinement by \citet{lin2021barf}, called Instant-NeRF, which is the method we use as our implementation base.\\
The Microsoft HoloLens is a multisensor platform, that, besides other sensor types, includes an RGB camera for the acquisition of image sequences, as well as an IMU that provides camera poses in real time. Besides images and corresponding camera poses, the camera parameters are the third component needed for a NeRF training. The fixed camera parameters can be obtained from the HoloLens directly without the need of calibration.\\
\section{Related Work}\label{sec:relatedwork}

NeRF emerged from the research field of novel view synthesis, where, based on a set of images with given poses depicting an object or a scene, convincing views for novel poses not included in the training set are to be rendered.
Early approaches in this field rely on image-based rendering \citep{zhang_chen_2004}, where new views are synthesized by blending and warping of the given posed images.
Most current research efforts however, aim at modeling the three-dimensional structure of the depicted scene based on the given views and using this 3D structure to render new views.
In doing so, space was initially modeled explicitly, for instance as voxels \citep{sitzmann_et_al_2019}.
Later, implicit representations emerged, modeling scenes three-dimensionally via the learned weights of fully-connected neural networks.
Similar to the implicit surface representations trained by 3D guidance, where surfaces are modeled as the zero crossing of a signed distance function \citep{park_et_al_2019b}, implicit surfaces can be used for view synthesis via neural rendering \citep{sitzmann_et_al_2019b}.\\
The break-through of NeRF \citep{mildenhall2020nerf}, however, came to pass by using an implicit radiance field consisting of position-dependent density and radiance which additionally depends on viewing direction.
This implicit field is queried and trained by differentiable volume rendering \citep{kajiya_von-herzen_1984}.
Along pixel-wise view rays, the radiance field is sampled and the resulting density and radiance values are integrated along the ray to determine the color value of the respective pixel.\\
One drawback of this original NeRF procedure is its inefficiency in terms of processing time with training times of ten hours and more and inference time of several minutes for the rendering of a single image.
Recently, current work has been proposed, which achieves significant improvements in rendering as well as training time \citep{chen_et_al_2022b, fridovich-keil_et_al_2022, mueller2022instantNgp}, with \citep{mueller2022instantNgp} reaching training times of few seconds for object scale scenes.\\
Another line of research focuses on improving the geometric quality of NeRFs, for instance by reverting back to implicit surface representations instead of density fields while retaining a NeRF-like rendering and training process \citep{wang_et_al_2021b} or even optimizing explicit triangle meshes with flexible topology \citep{munkberg_et_al_2021}.
Concerning geometric accuracy, various works have explored the stabilizing effect of incorporating depth cues into the 2D-image-based training process.
These can be derived from the training images themselves by sparse photogrammetric reconstruction \citep{deng_et_al_2022} or by means of active sensors such as LiDAR \citep{rematas_et_al_2022} or RGB-D cameras \citep{attal_et_al_2021}.\\
Furthermore, the reliance on camera poses for the training images has been reduced by refining or completely determining them during the training process \citep{jeong_et_al_2021, lin2021barf}.
Similarly, internal camera parameters \citep{jeong_et_al_2021} and even deblurring \citep{ma_et_al_2022} can be considered and determined as well.
\section{Methodology}\label{sec:methodology}
In order to move towards real-time 3D reconstruction using Instant-NeRFs, this section describes the single steps taken from image acquisition to obtaining 3D object models. At first, the data streaming pipeline is described. Thereafter certain prerequisites are introduced in order to include the images and poses into the training process, which represents the following step. At last, the fast 3D reconstruction from the trained Instant-NeRF is described.
\subsection{Data streaming}\label{sec:datastreaming}
In the following, the synchronized network connection between HoloLens (Server) and a PC (Client) for data streaming is described. The network is established through a Wifi connection.\\\\
\textbf{Server}\\\\
As previously mentioned, the HoloLens runs a server application, based on the implementation of \citet{dibene2022HoloLens}. Images of size 1920$\times$1080\,px can be obtained with frame rates of 15 or 30 frames per second (fps) from the server in the YUV color model with NV12 pixel representation \citep{microsoftYUV_NV12}. The image size is halved w.r.t. 24\,bit RGB representations, because every pixel is encoded with 12\,bits. This format has the least radiometric loss induced through compression we can obtain from the server. For purposes of a more rapid development and flexibility, we developed a server emulation tool that presents itself as a HoloLens in the network. The client application described below in standalone mode also provides the functionality to record streams of images and poses and save them persistently to the hard disk, so that such streams can be loaded into the server emulation tool. From there, we are able to change the framerate artificially, cut certain images out of the stream or only utilize a certain range of the image stream.\\\\
\textbf{Client}\\\\The client application is built directly into the Instant-NeRF implementation, which is programmed in C++ and Cuda C++, respectively. By sending instructions as byte arrays to the server, the start of streaming of images and poses is induced. The byte arrays include information about the required image size and framerate, besides other information. In addition before streaming, the camera parameters need to be queried from the server, which is achieved by sending a byte array that contains similar instructions as mentioned before. The byte order of the arrays needs to explicitly be converted to little endian representation before writing to the client socket.
\subsection{NeRF training}\label{sec:realtimetraining}
\begin{figure}
    \centering
    \includegraphics[width=0.40\textwidth]{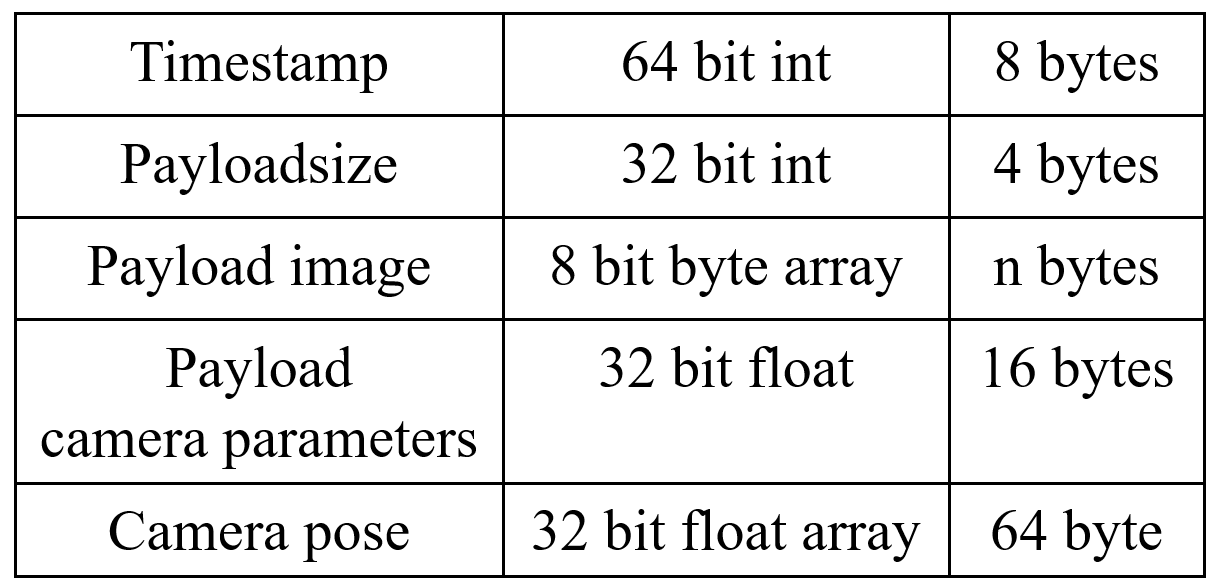}
    \caption{The structure of a streaming entity. On the left column, the content type is depicted, on the middle column the data type and on the right column the size in bytes. For the images that are streamed in our case, the payload image size is 1920$\times$1080$\times$1.5 bytes (12 bits), which results in 3110400 bytes.}
    \label{fig:entitystructure}
\end{figure}

This Section pays attention to the prerequisites and steps that need to be taken, in order to enable real-time training based on a stream of images and poses. Besides images and poses, also the camera parameters need to be taken into account and therefore transferred prior to the start of the training process.\\\\
\textbf{Camera parameters}\\\\After establishing a client-server connection between the HoloLens and PC, at first, the camera parameters need to be transferred. In the Instant-NeRF implementation the camera parameters can among others be represented in the OpenCV structure with five distortion coefficients in the form of $\{\,f_x, f_y, c_x, c_y, k_1, k_2, p_1, p_2, k_3\}$ \citep{opencv_cameracalibration}, where $f_x$ and $f_y$ denote the focal length, $c_x$ and $c_y$ the principal point, $k_i$ the radial symmetrical and $p_i$ the tangential asymmetrical distortion parameters. From the HoloLens, those parameters can be obtained either as single values or a 4$\times$4 camera matrix, represented as a 2D array. Both focal length parameters are provided in px by the HoloLens, which is the target unit for the OpenCV representation. The same goes for the principal point, which is however commonly provided as a [u, v] point in px units. Those camera parameters as well as the distortion parameters can be mapped directly to the OpenCV representation in Instant-NeRF.\\\\
\textbf{Pose transformation}\\\\
Due to photogrammetric methods like SfM, camera parameters and poses can be estimated from a set of images. For Instant-NeRF as well as other NeRF implementations, the open-source software COLMAP \citep{schoenberger2016sfm} is often utilized for this task. The estimation process is executed independently and only once for creating a set of input and ground truth data, so that the parameters and poses are known a priori to the NeRF training. In the case of the HoloLens, the parameters are known before and the poses are received while training.\\
Instant-NeRFs are trained within a 3D scene box, of which the extent and a scaling factor needs to be known in advance. These are calculated from all poses through a Python script that is part of the Instant-NeRF code base. As mentioned, we do not know the poses prior to the training, so the user needs to define a scene box through interaction with the HoloLens. We utilize the center of the box as our center of attention $P_c$ as well as a scaling factor $s$ that we take from the provided Python script. From empirical studies, we found that the directional portion of the HoloLens poses needs to be transformed in order to point into the general direction of $P_c$.\\\\
\textbf{Image buffer}\\\\Because the data stream can potentially be endless, but only finite memory for storing images is available, at one point in the streaming process previously acquired images need to be deallocated or replaced by newly acquired images. This could have the impact, that data is taken out of the training procedure that might still be relevant for reconstruction. Because we do not want to investigate such influence, we limit the buffer size so that 400 images can be received at maximum. For that matter, it needs to be noted that within the Instant-NeRF implementation, only pixel positions and not actual pixel values are selected randomly from the image set to construct rays through the scene for training.\\\\
\textbf{Training}\\\\The HoloLens, as well as our server emulation tool, sends a sequence of streaming entities with a certain data structure per entity, as depicted in Figure \ref{fig:entitystructure}. Each entity consists of a header composed of a time stamp as well as the payload size, which denotes the size of the entity in bytes. The following byte sequence contains the image in NV12 representation (see Section \ref{sec:datastreaming}), four camera parameters $f_x, f_y, c_x, c_y$ and the camera pose in the form of 16 32\,bit float values, that can be interpreted as a 4$\times$4 projection matrix. The four camera parameters are ignored, because we operate in fixed camera parameter mode. Instant-NeRF expects RGBA images, therefore the YUV NV12 image needs to be converted accordingly.
\\
As described in Section \ref{sec:realtimetraining} more specifically, GPU memory in the form of an image sequence buffer is allocated statically before training. In order to investigate the convergence behaviour of the network at streaming time, the number of images written into the buffer at once can be determined. This means that the client application waits for $n$ images to arrive, before writing them into the buffer. Similar to the image buffer, a pose buffer is allocated before training that allows for the same amount of poses as images. If $n$ is chosen to match the framerate, a block of $n$ images and poses are copied into the allocated image and pose buffers.\\
After sending the streaming instruction (see Section \ref{sec:datastreaming}, Client) to the HoloLens or server emulation tool, the stream of images and poses starts immediately. With the first $n_{sub}$ batch received, the start of the training is triggered. The training is executed until the last image is received, which, in our investigations, is at latest the point when the image buffer is saturated.

\subsection{NeRF 3D reconstruction}\label{sec:realtime3Dreconstruction}

In order to derive an explicit scene representation in the form of a dense point cloud with colors from the implicit radiance field, we sample rays from the training poses and capture the 3D coordinates where a respective ray reaches its first major density peak as is done in \citep{mueller2022instantNgp} for the rendering of depth maps. 
Instead of doing this per entire image, we randomly sample rays over all training images.
In sampling only from training image poses, we assume that this provides higher depth accuracy than sampling from interpolated test views not included in the training set.
This, of course, presupposes that the object of interest is continuously covered with training views.

\section{Experiments}\label{sec:experiments}

In this section, the experiments are laid out. While an extensive testing of the setup by streaming and parallel training was done in advance for arbitrary scenes, there was no possibility to directly stream the data from the HoloLens into our extended Instant-NeRF implementation for the specific scene we want to investigate. The object and environment the scene is composed of is described below in Section \ref{sec:objectandenvironment}, where we could not place our PC with hardware that is needed for training. Using a Laptop instead, we therefore recorded and saved a stream as described in Section \ref{sec:datastreaming}, which then was used in conjunction with our server emulation tool.
\subsection{Object and environment}\label{sec:objectandenvironment}
The objective is to reconstruct an industrial object, which, in our case, is a metal barrel with yellow paint \citep{haitz_et_al_2022}. With a cylindrical shape, its height is around 93\,cm and the diameter around 62\,cm. As can be seen, the barrel has some geometrical damage in the form of bumps, as well as texture in the form of corrosion and other radiometric variation. We chose a damaged barrel, because surface inspection of industrial objects, especially such barrels, is of ongoing interest for us. Further on in this work, we refer to the barrel as the object. The data recording is undertaken in an industrial environment, similar to a production facility, where the object is located in the middle of a quadratic area of approximately 10\,$m^2$. This yields enough space for moving in a circular trajectory around the object.
\subsection{Trajectory and data recording}\label{sec:trajectorydatastreaming}
\begin{figure}[t]
    \centering
    \subfigure[ ] {
        \includegraphics[width=0.35\textwidth]{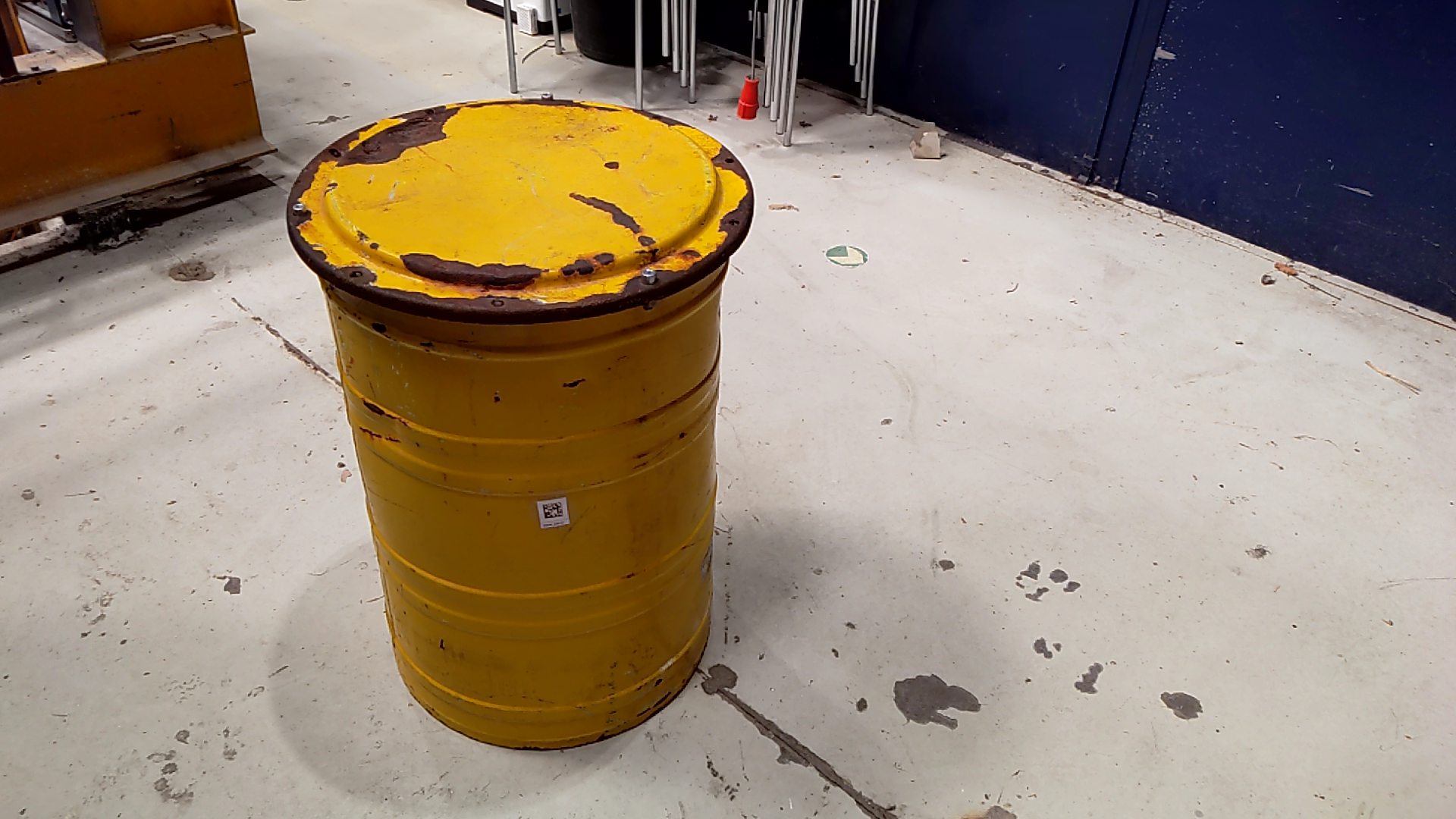}
      \label{fig:trajectoryimages_a}
    }
    \subfigure[ ] {
        \includegraphics[width=0.35\textwidth]{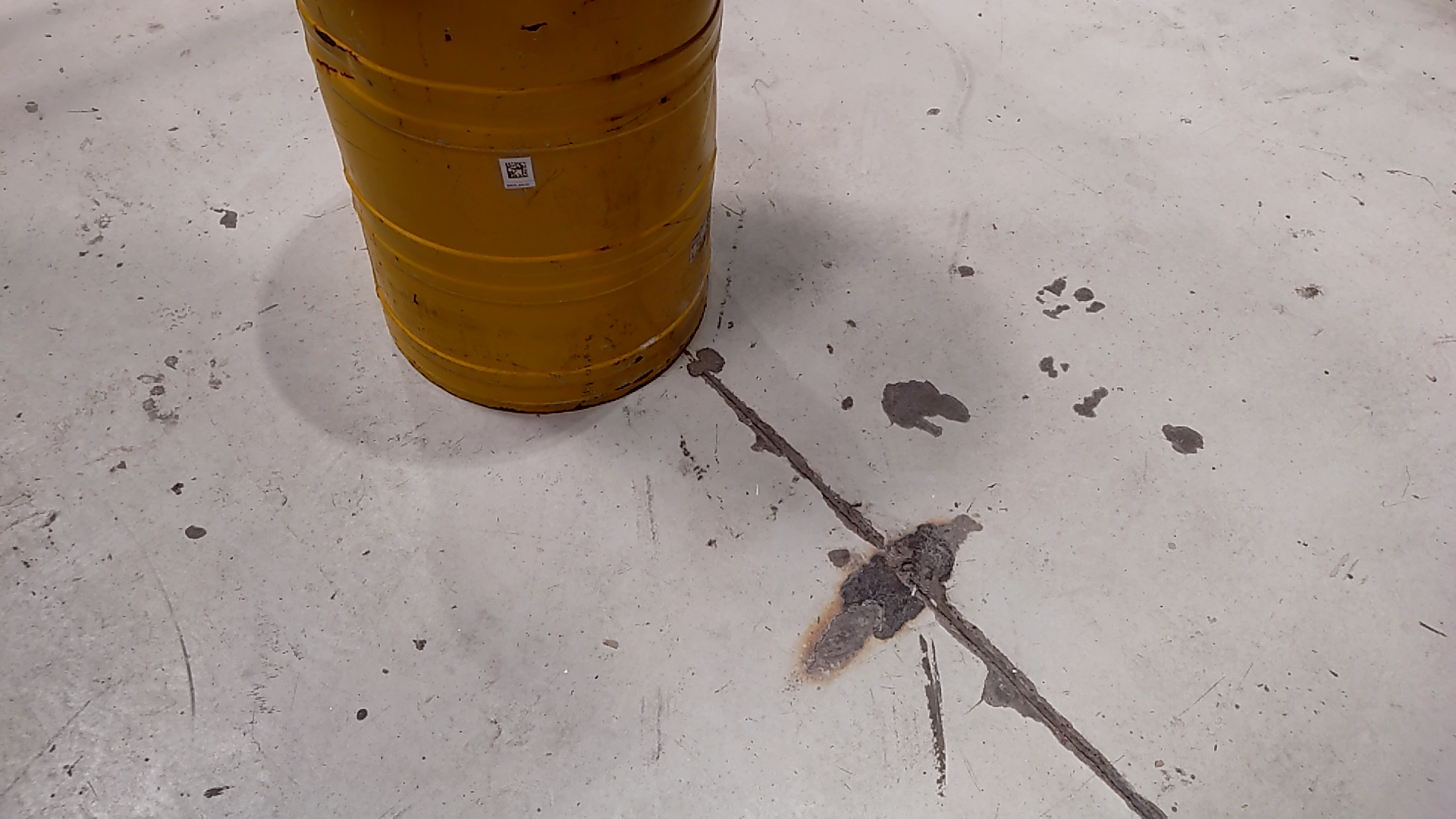}
      \label{fig:trajectoryimages_b}
    }
    \subfigure[ ] {
        \includegraphics[width=0.35\textwidth]{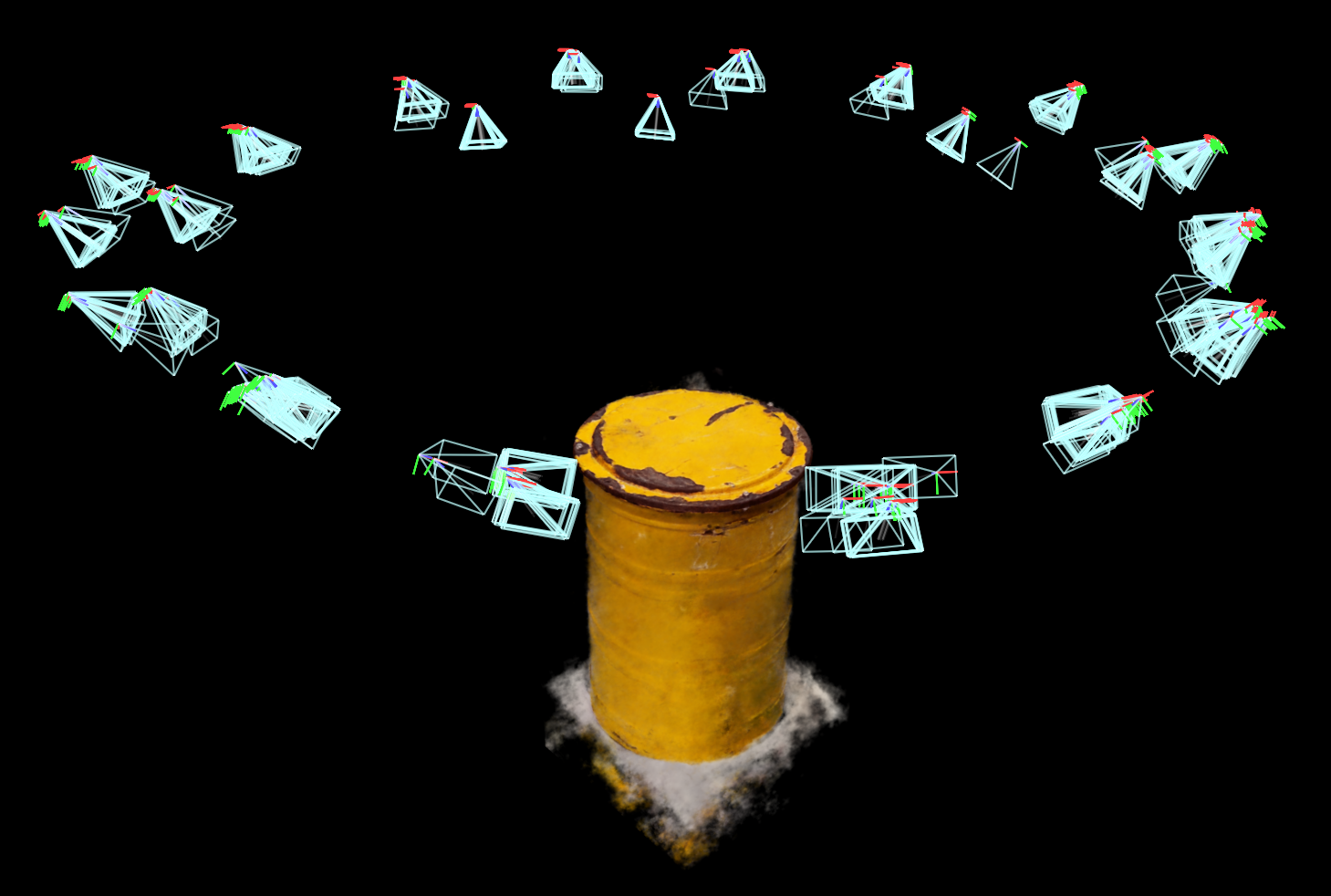}
      \label{fig:trajectoryimages_c}
    }
    \caption{
        Images regarding the trajectory. a) Image from first trajectory path, b) from second trajectory path. c) NeRF-based image reconstruction with visualized camera poses.
    }
    \label{fig:trajectoryimages}
\end{figure}

In order to record a data stream, a person wearing the HoloLens moves around the object on a circular path at least two times. Moving around the object for the first time, the person looks at the edge where the top and coat of the object meet (Figure \ref{fig:trajectoryimages_a}). On the second path, the person looks at the edge where the coat and the floor are meeting (Figure \ref{fig:trajectoryimages_b}). The coat needs to be visible in images of both paths in order to ensure the requirements for multi view consistency. The person takes one step at a time and does not move for a few seconds, so that images with a minimum of motion blur can be acquired by the camera. This can be observed in Figure \ref{fig:trajectoryimages_c}, where multiple poses heap up in certain spots. After the two paths are finished, the person moves in the same manner until the maximum number of 400 images (Section \ref{sec:realtimetraining}) is reached. For the whole recording process, the camera parameters are fixed.
\subsection{Training, reconstruction and hardware}\label{sec:trainingandreconstruction}
The training is realized using the server emulation tool, that listens at localhost in order to connect with the client in our modified Instant-NeRF implementation. After sending the corresponding instructions to the server, data streaming begins, which immediately triggers training. It needs to be noted that some recorded images have invalid poses, so that they need to be sorted out before at runtime. While all 400 images and poses are streamed, only 355 have valid poses which are used for training, whereas the remaining 45 images and poses are rejected. The server emulation tool allows for arbitrary framerates, so that the convergence behaviour of the NeRF can be investigated with different fps. The framerates 2\,fps, 5\,fps, 15\,fps and 30\,fps are investigated in particular, because 15\,fps and 30\,fps can be obtained through the HoloLens. With a low framerate of 5\,fps, the training time is prolonged, which might lead to better convergence on the one hand, and is also still a reasonable framerate for the scope of this work. In particular, 2\,fps need to be mentioned. This is the actual highest framerate that we could achieve with the HoloLens with the YUV NV12 image encoding using a Wifi connection.\\
Besides the Microsoft HoloLens, we utilize a PC for training and reconstruction which has an Intel i9 10850K CPU, 32\,GB Ram as well as a Nvidia Geforce RTX 3090 GPU, which is needed for training NeRFs efficiently.
\section{Results}\label{sec:results}
The results are categorized into qualitative and quantitative results, whereas for qualitative results image and geometric reconstructions are visualized. Image reconstruction is the ability of the NeRF to generate novel views. This is measured through the Peak-Signal-To-Noise-Ratio (PSNR) in the unit decibel (dB), which induces a logarithmic representation. Besides the reconstruction ability of NeRFs, the PSNR is utilized to measure the quality of lossy image compression algorithms. The geometric reconstruction is shown as point clouds, which is the target geometry type of this work. We have integrated a PLY-file writer into our modified Instant-NeRF implementation in order to save the point clouds to the hard disk. Writing point clouds to the hard disk is not part of the real-time processing and therefore is not added to the runtime investigation.\\
The runtime is measured from the training start, induced by the data stream, until the end of the 3D reconstruction. It needs to be noticed that the 3D reconstruction starts immediately after the training has been stopped. The termination criterion for training is the end of the data stream.\\
\subsection{Qualitative results}\label{sec:qualitativeresults}
\begin{figure*}
    \centering
    \subfigure[ ] {
        \includegraphics[width=0.15\textwidth]{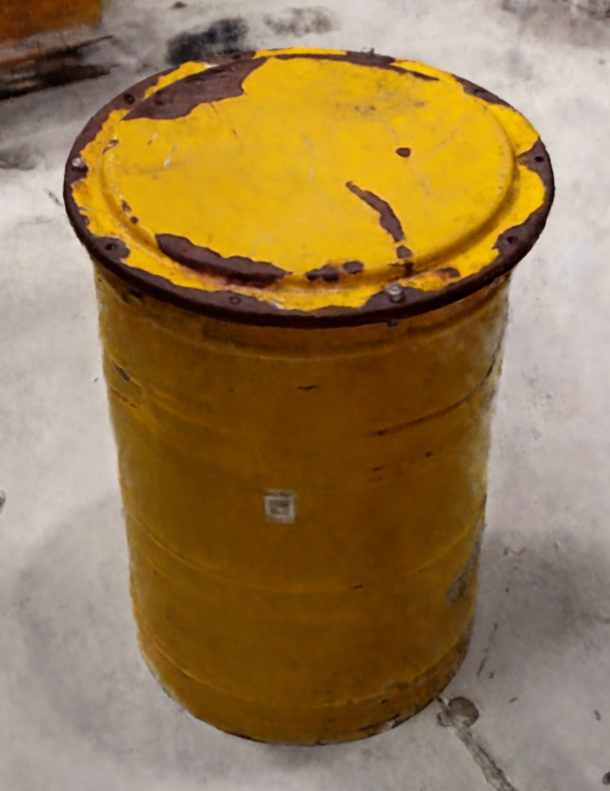}
      \label{fig:image_reconstruction_results_a}
    }
    \subfigure[ ] {
        \includegraphics[width=0.15\textwidth]{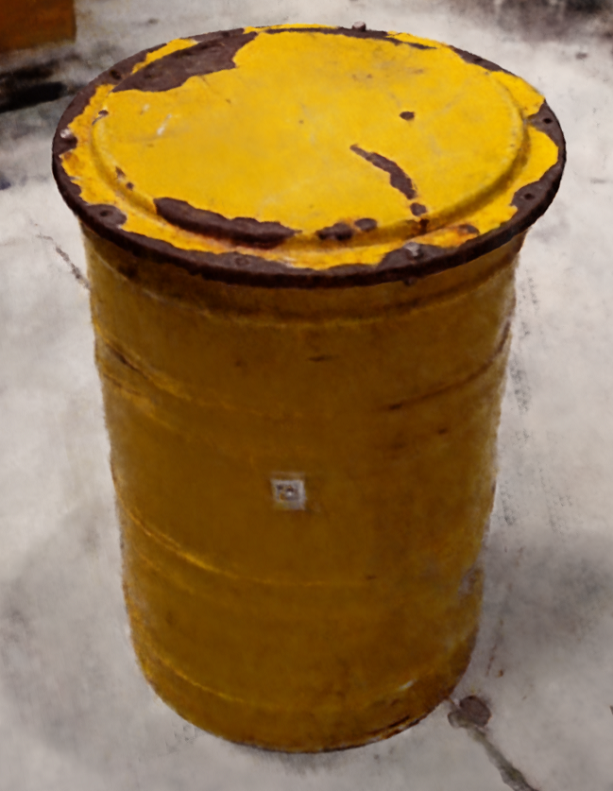}
      \label{fig:image_reconstruction_results_b}
    }
    \subfigure[ ] {
        \includegraphics[width=0.15\textwidth]{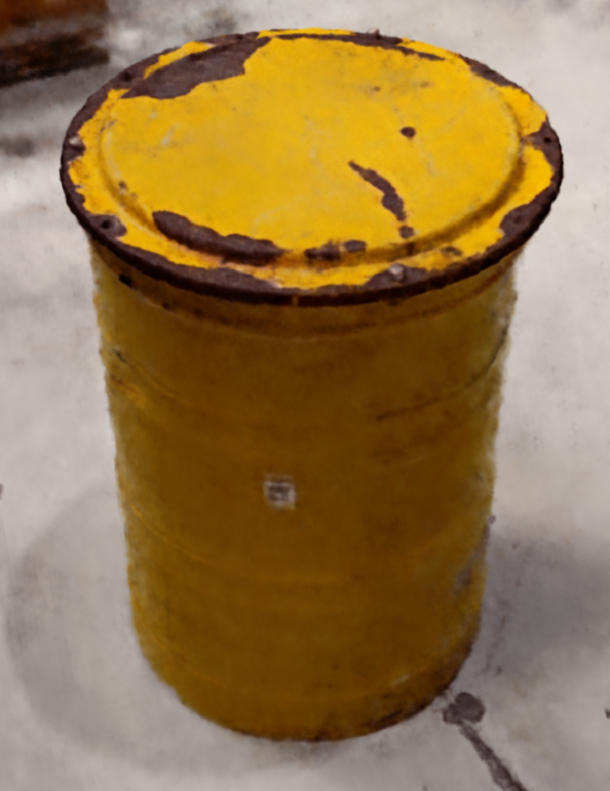}
      \label{fig:image_reconstruction_results_c}
    }
    \subfigure[ ] {
        \includegraphics[width=0.15\textwidth]{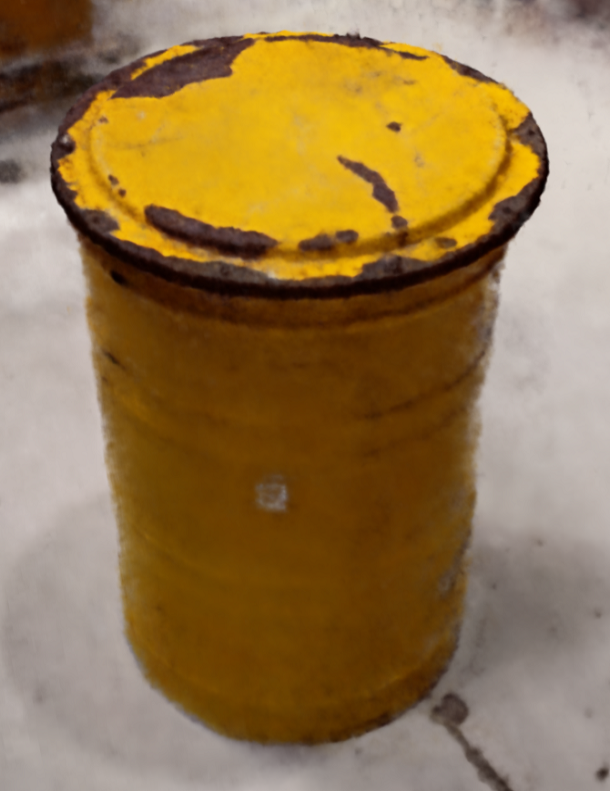}
      \label{fig:image_reconstruction_results_d}
    }
    \caption{
        Results for novel view image rendering for different training framerates a) 2\,fps b) 5\,fps c) 15\,fps d) 30\,fps.
    }
    \label{fig:image_reconstruction_results}
\end{figure*}

This section briefly describes the qualitative results of our experiments. The results are represented as images.\\\\
\textbf{Image reconstruction}\\\\
Figure \ref{fig:image_reconstruction_results} shows image reconstruction results rendered from a test pose not included in the training dataset for the different training framerates. A slight decreasing of visual quality is observable with increasing training framerate.\\\\

\begin{figure*}
    \centering
    \subfigure[] {
        \includegraphics[width=0.17\textwidth]{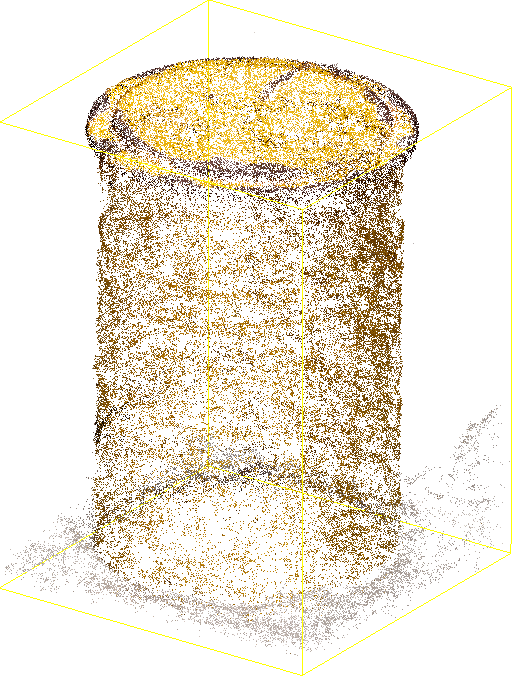}
        \label{fig:geometric_reconstruction_results_a}
    }
    \subfigure[ ] {
        \includegraphics[width=0.17\textwidth]{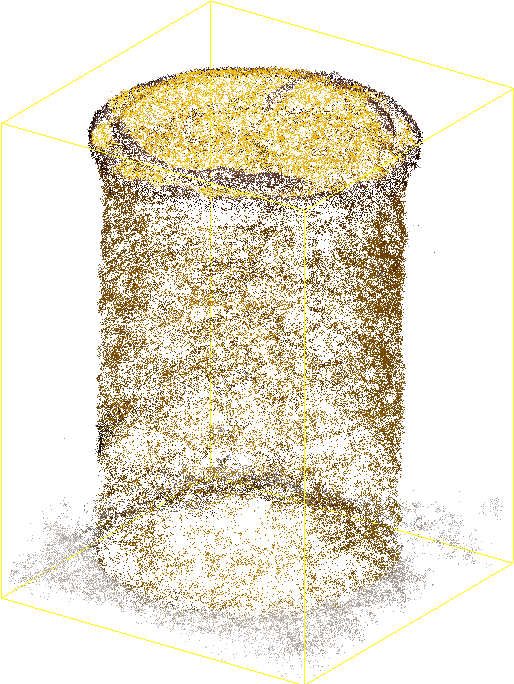}
        \label{fig:geometric_reconstruction_results_b}
    }
    \subfigure[ ] {
        \includegraphics[width=0.17\textwidth]{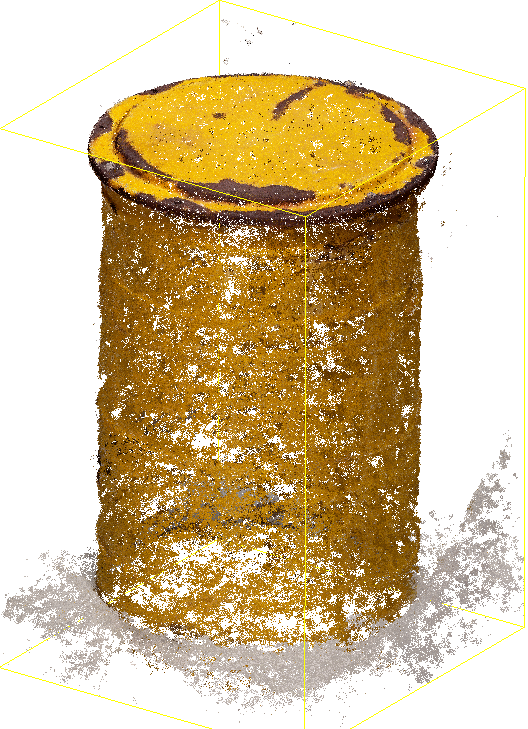}
        \label{fig:geometric_reconstruction_results_c}
    }
    \subfigure[ ] {
        \includegraphics[width=0.17\textwidth]{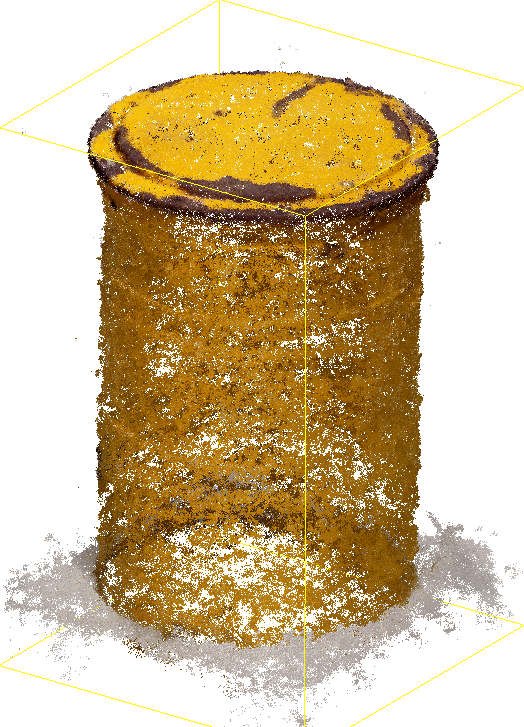}
        \label{fig:geometric_reconstruction_results_d}
    }
    \caption{
        Results for the reconstructed point clouds for different training framerates and different numbers of randomly sampled rays a) 2\,fps, 500k rays b) 30\,fps, 500k rays c) 2\,fps, 5000k rays d) 30\,fps, 5000k rays.
    }
    \label{fig:geometric_reconstruction_results}
\end{figure*}

\textbf{Geometric reconstruction}\\\\
Figure \ref{fig:geometric_reconstruction_results} examplarily shows geometric reconstruction results as point clouds sampled from the trained radiance fields. The point cloud density obviously increases with increasing number of sampled rays. The differences between training framerates of 2\,fps and 30\,fps for a given number of sampled rays is however visually hardly noticeable.

\subsection{Quantitative results}\label{sec:quantitativeresults}
In Table \ref{tab:performance5000k}, runtime performance and image reconstruction ability is shown. All things considered, 400 images are streamed from the server emulation tool into our modified Instant-NeRF application, of which 355 are used for training. A dense reconstruction with five million (5000k) scene points is executed in this investigation. Table \ref{tab:performancesparsedense} depicts the performance of the 3D reconstruction dependent on the target number of points that need to be extracted. Lastly, Table \ref{tab:extractedpoints} shows the actual extracted points w.r.t. the target number of points.
\begin{table}[h]
    \centering
    \begin{tabular}{c|ccccc}
    \multicolumn{1}{c|} {FR}&      Train&     3DR&      ACC&       PSNR\\ 
                    fps     & s         & s         & s         & dB \\ \hline
                    -       & 600.0     & 1.08     & 601.08   & 29.98 \\
                    2       & 198.906   & 0.788     & 199.694   & 27.85 \\
                    5       & 81.892    & 0.862     & 82.754    & 26.98 \\
                    15      & 30.404    & 0.986     & 31.390    & 26.04 \\
                    30      & 19.042    & 1.115     & 20.139    & 25.68 
    \end{tabular}
    \caption{Runtime performance and reconstruction ability. Framerate (FR) is shown in frames per second (fps), Instant-NeRF training time (Train), 3D reconstruction time (3DR) and both accumulated (ACC) are depicted in seconds (s). The reconstruction ability is shown as the peak signal-to-noise-ratio in decibel (dB).}
    \label{tab:performance5000k}
\end{table}

\begin{table}[h]
    \centering
    \begin{tabular}{c|ccccc}
    \multicolumn{1}{c|} {FR}&      500k&     5000k&\\ 
                    fps     & s         & s     \\ \hline
                    2       & 0.079   & 0.788 \\
                    5       & 0.085    & 0.862\\
                    15      & 0.101    & 0.986\\
                    30      & 0.116    & 1.115
    \end{tabular}
    \caption{Reconstruction performance for 500k and 5000k points to be extracted after training.}
    \label{tab:performancesparsedense}
\end{table}

\begin{table}[h]
    \centering
    \begin{tabular}{c|ccccc}
    \multicolumn{1}{c|} {FR}& 5000k & PSNR \\ 
                    fps       & points   & dB \\ \hline
                    2         & 4659291 & 27.85\\
                    5       & 4680735 & 26.98\\
                    15          & 4920716 & 26.04\\
                    30         & 4967545 & 25.68
    \end{tabular}
    \caption{Framerate, number of hit points within the scene and the corresponding PSNR at training end.}
    \label{tab:extractedpoints}
\end{table}

\section{Discussion}\label{sec:discussion}
In this section, the results are discussed in the structure of image reconstruction, geometric reconstruction and runtime performance. The latter includes a discussion of the runtime of training and reconstruction, each.
\subsection{Image reconstruction}\label{sec:discussionimage}
Rendering new views of a scene is the task NeRFs are commonly used for. While image rendering is less relevant for computer vision tasks, some insights can be derived from the experiments. As shown quantitatively in Table \ref{tab:performance5000k}, the PSNR increases with training time, which is expected if training image quality and pose accuracy is sufficient in cases where the data is available before training (i.e. not real-time). For real-time training based on a data stream with a finite image buffering, however, the framerate used for streaming is an important factor, as it determines the training time in this setup. In case of no real-time training and when the data is available before training, the data row of Table \ref{tab:performance5000k} shows that the PSNR begins to peak after 10 minutes of training with 29.98\,dB.\\
The qualitative results in Figure \ref{fig:image_reconstruction_results} show no significant difference regarding the object itself. Different PSNR values rather indicate differences in the background reconstruction ability and the occurrence of noise and artifacts.
\subsection{Geometric reconstruction}\label{sec:discussiongeometric}
For the geometric reconstruction some very interesting insights can be obtained from the experiments. At first, in Table \ref{tab:extractedpoints} the actual number of extracted points can be seen. While a number of points to be extracted is given with 5000k, the reconstruction algorithm samples rays that might not terminate at a point within the scene because of low density throughout its path. Consequently, no point will be added to the point cloud for the sampled ray. Also Table \ref{tab:extractedpoints} shows that with increasing PSNR (i.e. image reconstruction ability), the number of points decreases. This is very likely due to less noise, which is indicated through high PSNR. Besides that, common artifacts like floaters are less present with better image reconstruction. Floaters in the NeRF terminology are geometric anomalies, very likely introduced by inaccurate camera parameters and poses \citep{barron2022mipnerf360, sabour2023robustnerf}. If less noise and artifacts are present in the scene, there is a slightly higher probability that the ray does not terminate within the scene, which therefore might result in a lower number of points. Both noise and artifacts however can also lead to early ray termination, which lead to false geometric representations within the scene.\\
The qualitative results are shown with a more sparse (500k points) and a more dense (5000k) point cloud in Figure \ref{fig:geometric_reconstruction_results}. Similar to the image reconstruction, with lower PSNR the number of noise and artifacts appears to increase. This can be seen i.e. in Figure \ref{fig:geometric_reconstruction_results_c} on the lower right (floater-like artifacts) and above the barrel (noise).
\subsection{Runtime performance}\label{sec:discussionruntime}
The runtime performance can be split into training and reconstruction. While the training time is determined by the start and end of the data stream, there is a relatively small overhead due to implementation details. As an example in Table \ref{tab:performance5000k}, for a framerate of 5 fps the training time takes 81.892 seconds. If 400 images are streamed as a whole, the streaming time should be 80 seconds for 5 fps. \\Based on massive parallel processing using the Cuda framework and a dedicated GPU (see Section \ref{sec:trainingandreconstruction}), we are able to write around 5000k scene points as a dense point cloud into CPU Ram in around 1 second as shown in Tables \ref{tab:performance5000k} and \ref{tab:performancesparsedense} for the geometric reconstruction. Most experiments were based on the target number of 5000k points, for comparison, however, we also set the target number to 500k, which results in a more sparse point cloud. The runtime results comparing both target numbers are shown in Table \ref{tab:performancesparsedense}. For the four shown scenarios, the connection of runtime performance and number of extracted points appears to be largely linear. 
\section{Conclusion}\label{sec:conclusion}
In this work, we described a setup and an implementation for real-time NeRF training and instant geometric 3D reconstruction. Some compromises had to be made in order for the training and reconstruction to work. At first, a manual scene box needs to be annotated by the user in interaction with the HoloLens. Due to the implementation of Instant-NeRF, such a box needs to be known or determined in advance. Usually when the data is at hand before training, the scene box is derived from the camera poses. Another challenge is the image buffer, which we initialized statically before runtime. The reason for this was to keep all acquired images and poses for training and not filter out images. The latter would have caused that training data would not have been around for very long, which might impact the reconstruction quality negatively. However, this is a very interesting study area, because it is of practical use to stream more images than the GPU memory size allows and to find a strategy, that keeps only the images for good reconstruction ability. In case of the HoloLens, we investigate a format with a higher level of compression in order to stream images with 15\,fps or 30\,fps opposed to the only 2\,fps we could obtain with the YUV NV12 format.\\
Besides ongoing challenges of this work, we obtained exceptional results regarding the runtime performance. To be able to geometrically reconstruct a scene with 5000k points within one second after a data stream has high potential for real-time 3D mobile mapping. Regarding the hardware, a high performance laptop with an Nvidia GPU can be utilized. Depending on the GPU RAM, which usually ranges between 8 and 16 Gb RAM for mobile GPUs, the image buffer size as well as the batch size determine how much GPU RAM is used. In our experiments, we did not exceed 7 Gb GPU RAM for training and reconstruction.\\
Regarding the applications, we aim towards geometric damage detection in industrial objects. But because we are also interested in radiometric damage, the integration of 2D semantic segmentation as carried out in \citet{haitz_et_al_2022b} is a planned task. The utilization of knowledge distillation into the NeRF \citep{kobayashi2022distilledfeaturefields} is therefore of high interest for us.
{
	\begin{spacing}{1.17}
		\normalsize
		\bibliography{literature} 

\begin{thebibliography}{xx}

\bibitem[Attal et al., 2021]{attal_et_al_2021}
Attal, B., Laidlaw, E., Gokaslan, A., Kim, C., Richardt, C., Tompkin, J.,
  O'Toole, M., 2021.
 {TöRF: Time-of-Flight Radiance Fields for Dynamic Scene View Synthesis}.
 \emph{Advances in Neural Information Processing Systems (NeurIPS)}, ~34,
  1--13.

\bibitem[Barron et al., 2022]{barron2022mipnerf360}
Barron, J.~T., Mildenhall, B., Verbin, D., Srinivasan, P.~P., Hedman, P., 2022.
 Mip-NeRF 360: Unbounded Anti-Aliased Neural Radiance Fields.
 {\em CVPR}.

\bibitem[Chen et al., 2022]{chen_et_al_2022b}
Chen, A., Xu, Z., Geiger, A., Yu, J., Su, H., 2022.
 {TensoRF: Tensorial Radiance Fields}.
 {\em arXiv}, 2203.09517, 1-17.

\bibitem[Dellaert and Yen-Chen, 2020]{dellart_yen-chen_2020}
Dellaert, F., Yen-Chen, L., 2020.
 {Neural Volume Rendering: NeRF And Beyond}.
 {\em arXiv}, 2101.05204, 1-8.

\bibitem[Deng et al., 2022]{deng_et_al_2022}
Deng, K., Liu, A., Zhu, J.-Y., Ramanan, D., 2022.
 {Depth-Supervised NeRF: Fewer Views and Faster Training for Free}.
 \emph{IEEE/CVF Conference on Computer Vision and Pattern Recognition (CVPR)},
  12882--12891.

\bibitem[Dibene and Dunn, 2022]{dibene2022HoloLens}
Dibene, J.~C., Dunn, E., 2022.
 HoloLens 2 Sensor Streaming.
 {\em arXiv preprint arXiv:2211.02648}.

\bibitem[Dubois et al., 2021]{dubiosjutzi2021}
Dubois, C., Jutzi, B., Olijslagers, M., Pathe, C., Schmullius, C.,
  Stelmaszczuk-G\'orska, M.~A., Vandenbroucke, D., Weinmann, M., 2021.
 Knowledge And Skills Related To Active Optical Sensors In The Body Of
  Knowledge For Earth Observation And Geoinformation (EO4GEO BOK).
 {\em ISPRS Annals of the Photogrammetry, Remote Sensing and Spatial
  Information Sciences}, V-5-2021, 9--16.
 https://www.isprs-ann-photogramm-remote-sens-spatial-inf-sci.net/V-5-2021/9/2021/.

\bibitem[Fridovich-Keil et al., 2022]{fridovich-keil_et_al_2022}
Fridovich-Keil, S., Yu, A., Tancik, M., Chen, Q., Recht, B., Kanazawa, A.,
  2022.
 {Plenoxels: Radiance Fields Without Neural Networks}.
 \emph{IEEE/CVF Conference on Computer Vision and Pattern Recognition (CVPR)},
  5501--5510.

\bibitem[Gao et al., 2022]{gao_et_al_2022}
Gao, K., Gao, Y., He, H., Lu, D., Xu, L., Li, J., 2022.
 {NeRF: Neural Radiance Field in 3D Vision, A Comprehensive Review}.
 {\em arXiv}, 2210.00379, 1-20.

\bibitem[Haitz et al., 2022a]{haitz_et_al_2022b}
Haitz, D., Hübner, P., Ulrich, M., Landgraf, S., Jutzi, B., 2022a.
 {Semantic Segmentation with Small Training Datasets: A Case Study for
  Corrosion Detection on the Surface of Industrial Objects}.
 \emph{Image Processing Forum}, 73--85.

\bibitem[Haitz et al., 2022b]{haitz_et_al_2022}
Haitz, D., Jutzi, B., Hübner, P., Ulrich, M., 2022b.
 {Corrosion Detection for Industrial Objects: From Multi-Sensor System to 5D
  Feature Space}.
 {\em The International Archives of the Photogrammetry, Remote Sensing and
  Spatial Information Sciences}, XLIII-B1-2022, 143-150.

\bibitem[Hartley and Zisserman, 2004]{Hartley2004}
Hartley, R.~I., Zisserman, A., 2004.
 {\em Multiple View Geometry in Computer Vision}.
 Second edn, Cambridge University Press.

\bibitem[H{\"{u}}bner et al., 2020]{huebner_et_al_2020}
H{\"{u}}bner, P., Clintworth, K., Liu, Q., Weinmann, M., Wursthorn, S., 2020.
 {Evaluation of HoloLens Tracking and Depth Sensing for Indoor Mapping
  Applications}.
 {\em Sensors}, 20(4), 1021:1-24.

\bibitem[J{\"a}ger et al., 2023]{jaeger_et_al_2023}
J{\"a}ger, M., H{\"u}bner, P., Haitz, D., Jutzi, B., 2023.
 {A Comparative Neural Radiance Field (NeRF) 3D Analysis of Camera Poses from
  HoloLens Trajectories and Structure from Motion}.
 {\em arXiv}, 2304.10664, 1-7.

\bibitem[Jeong et al., 2021]{jeong_et_al_2021}
Jeong, Y., Ahn, S., Choy, C., Anandkumar, A., Cho, M., Park, J., 2021.
 {Self-Calibrating Neural Radiance Fields}.
 \emph{IEEE/CVF International Conference on Computer Vision (ICCV)},
  5846--5854.

\bibitem[Jutzi et al., 2015]{JutziMeyerHinz2015_1000070460}
Jutzi, B., Meyer, F., Hinz, S., 2015.
 {\em Aktive Fernerkundungssensorik – Technologische Grundlagen und
  Abbildungsgeometrie}.
 {Springer Verlag}, 1–40.

\bibitem[Kajiya and {Von Herzen}, 1984]{kajiya_von-herzen_1984}
Kajiya, J.~T., {Von Herzen}, B.~P., 1984.
 {Ray Tracing Volume Densities}.
 {\em ACM SIGGRAPH Computer Graphics}, 18(3), 165-174.

\bibitem[Kobayashi et al., 2022]{kobayashi2022distilledfeaturefields}
Kobayashi, S., Matsumoto, E., Sitzmann, V., 2022.
 Decomposing nerf for editing via feature field distillation.
 \emph{Advances in Neural Information Processing Systems}, ~35.

\bibitem[Lin et al., 2021]{lin2021barf}
Lin, C.-H., Ma, W.-C., Torralba, A., Lucey, S., 2021.
 Barf: Bundle-adjusting neural radiance fields.
 \emph{IEEE International Conference on Computer Vision ({ICCV})}.

\bibitem[Ma et al., 2022]{ma_et_al_2022}
Ma, L., Li, X., Liao, J., Zhang, Q., Wang, X., Wang, J., Sander, P.~V., 2022.
 {Deblur-NeRF: Neural Radiance Fields from Blurry Images}.
 {\em IEEE/CVF Conference on Computer Vision and Pattern Recognition (CVPR)},
  12861-1287.

\bibitem[Microsoft, 2023]{microsoftYUV_NV12}
Microsoft, 2023.
 Microsoft app development.
 Accessed: 2023-04-19.

\bibitem[Mildenhall et al., 2020]{mildenhall2020nerf}
Mildenhall, B., Srinivasan, P.~P., Tancik, M., Barron, J.~T., Ramamoorthi, R.,
  Ng, R., 2020.
 Nerf: Representing scenes as neural radiance fields for view synthesis.
 \emph{ECCV}.

\bibitem[M\"uller et al., 2022]{mueller2022instantNgp}
M\"uller, T., Evans, A., Schied, C., Keller, A., 2022.
 Instant Neural Graphics Primitives with a Multiresolution Hash Encoding.
 {\em ACM Trans. Graph.}, 41(4), 102:1--102:15.
 https://doi.org/10.1145/3528223.3530127.

\bibitem[Munkberg et al., 2022]{munkberg_et_al_2021}
Munkberg, J., Hasselgren, J., Shen, T., Gao, J., Chen, W., Evans, A., Müller,
  T., Fidler, S., 2022.
 {Extracting Triangular 3D Models, Materials, and Lighting From Images}.
 \emph{IEEE/CVF Conference on Computer Vision and Pattern Recognition (CVPR)},
  8280--8290.

\bibitem[Müller et al., 2022]{muellerInstantNerf}
Müller, T., Evans, A., Schied, C., Foco, M., B\'{o}dis-Szomor\'{u}, A.,
  Deutsch, I., Shelley, M., Keller, A., 2022.
 Instant neural radiance fields.
 \emph{ACM SIGGRAPH 2022 Real-Time Live!}, SIGGRAPH '22, Association for
  Computing Machinery, New York, NY, USA.

\bibitem[OpenCV, 2023]{opencv_cameracalibration}
OpenCV, 2023.
 Opencv: Camera calibration.
 Accessed: 2023-04-19.

\bibitem[Park et al., 2019]{park_et_al_2019b}
Park, J.~J., Florence, P., Straub, J., Newcombe, R., Lovegrove, S., 2019.
 {DeepSDF: Learning Continuous Signed Distance Functions for Shape
  Representation}.
 \emph{IEEE/CVF Conference on Computer Vision and Pattern Recognition (CVPR)},
  165--174.

\bibitem[Rematas et al., 2022]{rematas_et_al_2022}
Rematas, K., Liu, A., Srinivasan, P.~P., Barron, J.~T., Tagliasacchi, A.,
  Funkhouser, T., Ferrari, V., 2022.
 {Urban Radiance Fields}.
 \emph{IEEE/CVF Conference on Computer Vision and Pattern Recognition (CVPR)},
  12932--12942.

\bibitem[Rosenblatt, 1961]{mlp-rosenblatt1961principles}
Rosenblatt, F., 1961.
 Principles of neurodynamics. perceptrons and the theory of brain mechanisms.
 Technical report, Cornell Aeronautical Lab Inc Buffalo NY.

\bibitem[Sabour et al., 2023]{sabour2023robustnerf}
Sabour, S., Vora, S., Duckworth, D., Krasin, I., Fleet, D.~J., Tagliasacchi,
  A., 2023.
 Robustnerf: Ignoring distractors with robust losses.

\bibitem[Sch\"{o}nberger and Frahm, 2016]{schoenberger2016sfm}
Sch\"{o}nberger, J.~L., Frahm, J.-M., 2016.
 Structure-from-motion revisited.
 \emph{Conference on Computer Vision and Pattern Recognition (CVPR)}.

\bibitem[Sitzmann et al., 2019a]{sitzmann_et_al_2019}
Sitzmann, V., Thies, J., Heide, F., Nießner, M., Zollhöfer, G. W.~M., 2019a.
 {DeepVoxels: Learning Persistent 3D Feature Embeddings}.
 \emph{IEEE/CVF Conference on Computer Vision and Pattern Recognition (CVPR)},
  ~1, 2432--2441.

\bibitem[Sitzmann et al., 2019b]{sitzmann_et_al_2019b}
Sitzmann, V., Zollhoefer, M., Wetzstein, G., 2019b.
 {Scene Representation Networks: Continuous 3D-Structure-Aware Neural Scene
  Representations}.
 \emph{Advances in Neural Information Processing Systems}, ~32, 1--12.

\bibitem[Steger et al., 2018]{StegerUlrichWiedemann2018}
Steger, C., Ulrich, M., Wiedemann, C., 2018.
 {\em Machine Vision Algorithms and Applications}.
 2nd edn, {Wiley-VCH Verlag}.

\bibitem[Wang et al., 2021]{wang_et_al_2021b}
Wang, P., Liu, L., Liu, Y., Theobalt, C., Komura, T., Wang, W., 2021.
 {NeuS: Learning Neural Implicit Surfaces by Volume Rendering for Multi-view
  Reconstruction}.
 \emph{Advances in Neural Information Processing Systems (NeurIPS)},  34
  (pre-proceedings), 1--13.

\bibitem[Zhang and Chen, 2004]{zhang_chen_2004}
Zhang, C., Chen, T., 2004.
 {A Survey on Image-Based Rendering - Representation, Sampling and
  Compression}.
 {\em Signal Processing: Image Communication}, 19(1), 1-28.

\end{thebibliography}
	\end{spacing}
}

\end{document}